\documentclass[12pt]{iopart}

\usepackage{iopams}
\usepackage{amstext}

\usepackage{graphicx}
\usepackage{subcaption}
\usepackage{multirow}

\usepackage{appendix}

\usepackage{booktabs}
\usepackage{threeparttable}
\usepackage{makecell} % Add this for advanced cell formatting
\newcommand{\cnxblock}[3]{
  \begin{minipage}[c]{0.27\textwidth} 
    \centering
    \vspace{1.5mm} 
    $\left[ \begin{array}{rl}
      \mbox{d1} \times 7, & #1 W \\
      1 \times 1, & #2 W \\
      1 \times 1, & #1 W \\
    \end{array} \right] \times #3 D$
  \vspace{1.5mm} % Adds space at the bottom
\end{minipage}
} 

\usepackage[numbers,compress]{natbib}
\usepackage{setspace}
\usepackage[hidelinks]{hyperref}
\hypersetup{
    colorlinks=true,
    linkcolor=red, % Color of internal links
    urlcolor=blue,  % Color of external links
    citecolor=blue % Color of citations
}
\begin{document}

\title[Scaling CNNs achieves expert-level EEG seizure detection]{Scaling convolutional neural networks achieves expert-level seizure detection in neonatal EEG}

\author{Robert Hogan$^1$, Sean R. Mathieson$^{1,2}$, Aurel Luca$^1$, Soraia Ventura$^{1,2,3}$, Sean Griffin$^1$, Geraldine B. Boylan$^{1,2,3}$, and John M. O'Toole$^1$} 

\address{$^1$Cergenx Ltd, Dublin, Ireland\\
  $^2$INFANT Research Centre, University College Cork, Ireland\\
  $^3$Department of Paediatrics and Child Health, University College Cork, Ireland}
\ead{rhogan@cergenx.com, jotoole@cergenx.com}
\vspace{10pt}
\begin{indented}
\item[]May 2024
\end{indented}

\begin{abstract}
  {\it Background:} Neonatal seizures are a neurological emergency that require urgent
treatment. They are hard to diagnose clinically and can go undetected if EEG monitoring is
unavailable. EEG interpretation requires specialised expertise which is not widely available. Algorithms to detect EEG seizures can address this limitation but have yet to reach widespread clinical adoption.
  
  {\it Methods:} 
  Retrospective EEG data from 332 neonates was used to develop and validate a seizure-detection model.
  The model was trained and tested with a development dataset ($n=202$) that was annotated with over 12~k seizure events on a per-channel basis.
  This dataset was used to develop a convolutional neural network (CNN) using a modern architecture and training methods. 
  The final model was then validated on two independent multi-reviewer datasets ($n=51$ and $n=79$).
  
  {\it Results:} 
  Increasing dataset and model size improved model performance: Matthews correlation coefficient (MCC) and Pearson's correlation ($r$) increased by up to 50\% with data scaling and up to 15\% with model scaling.
  Over 50~k hours of annotated single-channel EEG was used for training a model with 21 million parameters.  
  State-of-the-art was achieved on an open-access dataset (MCC=0.764, $r=0.824$, and AUC=0.982). 
  The CNN attains expert-level performance on both held-out validation sets, with no significant difference in inter-rater agreement among the experts and among experts and algorithm ($\Delta \kappa < -0.095$, $p>0.05$). 
  
  {\it Conclusion:} With orders of magnitude increases in data and model scale we have produced a new state-of-the-art model for neonatal seizure detection.
  Expert-level equivalence on completely unseen data, a first in this field, provides a strong indication that the model is ready for further clinical validation.  

\end{abstract}

\section{Introduction}

The most frequent cause of neonatal seizures is acute brain injury in the early postnatal period.
Seizures typically emerge over the first 72 postnatal hours in term neonates, primarily caused by hypoxic-ischaemic encephalopathy (HIE) or cerebrovascular injury \cite{Tekgul2006,Soul2018,Pisani2021}. More than half of neonates with moderate or severe HIE develop seizures \cite{Tekgul2006,Rennie2018,Pisani2021}. 
For those neonates who do develop seizures, approximately 7\% to 10\% are at risk of death and  23\% to 50\% are at risk of poor outcome \cite{UriaAvellanal2013,Tekgul2006,Srinivasakumar2015}.

Seizures can be subtle, often without clinical correlate, and often remain undetected \cite{Murray2008}. 
Continuous electroencephalogram (EEG) monitoring is the gold standard for neonatal seizure surveillance. 
Yet real-time interpretation of the EEG requires specialised expertise that is not always available, limiting the capacity for continuous review of EEGs.
A recent multi-centre study found that, even with continuous EEG or amplitude-integrated EEG readily available, only 11\% of seizures were treated within 1-hour of onset \cite{Rennie2018}.
Prompt treatment can reduce seizure burden and therefore may reduce seizure-mediated neuronal damage and improve outcomes \cite{Pavel2022a}. 

Automated review of the EEG, with expert oversight, would allow for increased monitoring of at-risk neonates. A recent clinical trial of an automated algorithm to detect EEG seizures demonstrated the potential clinical utility \cite{Pavel2020}. Yet this seizure-detection algorithm, which was developed in 2011 \cite{Temko2011}, has been comprehensively surpassed in performance by a range of newer methods \cite{Ansari2019,pmlr-v126-isaev20a,OShea2020,Daly2021,Caliskan2021,Tanveer2021,Gramacki2022,Raeisi2022,Daly2023,Raeisi2023}.

Most contemporary seizure-detection methods use deep neural networks. These methods tend to improve with more training data, often beyond the limits of a machine-learning approach using pre-defined feature sets. 
A key to this success has been scaling both the neural-network architecture and the training set. To date, this scaling approach has not been applied to neonatal EEG, mostly due to the paucity of large datasets and the high cost of training larger models. Our primary goal in this study is to develop an algorithm to detect seizures in neonatal EEG with a high-level of accuracy suitable for clinical adoption. As part of this goal, we aim to test the hypothesis that increasing model size and training data will lead to better performance.

\section{Methods}

\subsection{Dataset Description}
\label{sec:dataset_details}

A development dataset is used for training and testing of the models, with a random 80:20 split. When the final models are fixed, which includes training on 100\% of the development set, they are then validated on two independent held-out datasets.

\subsubsection{Development Dataset}
We used a fully anonymized dataset of EEG recordings from 202 term neonates. That dataset, which we refer to as the Neobase, was collected at Cork University Maternity Hospital (CUMH), Ireland, over a 19 year time period as part of ongoing clinical research studies. 
% \cite{Murray2008,Greene2008,Temko2011,OShea2020,Garvey2022,Pavel2020,Pavel2023,Pavel2024,Korotchikova2009,Korotchikova2016,Raurale2020,Garvey2021}
Demographic and clinical data, presented in Table~\ref{tab:clin_demographics}, indicates a diverse cohort of neonates including EEG recording in the neonatal intensive care unit (NICU) from neonates with varied clinical indications (HIE, birth asphyxia, stroke) in addition to EEG for healthy controls recorded in the postnatal ward. Appendix~\ref{supp:dev_dataset} includes a more detailed description of dataset properties and collection methods.

\begin{figure}
    \centering
    \includegraphics[width=1\linewidth]{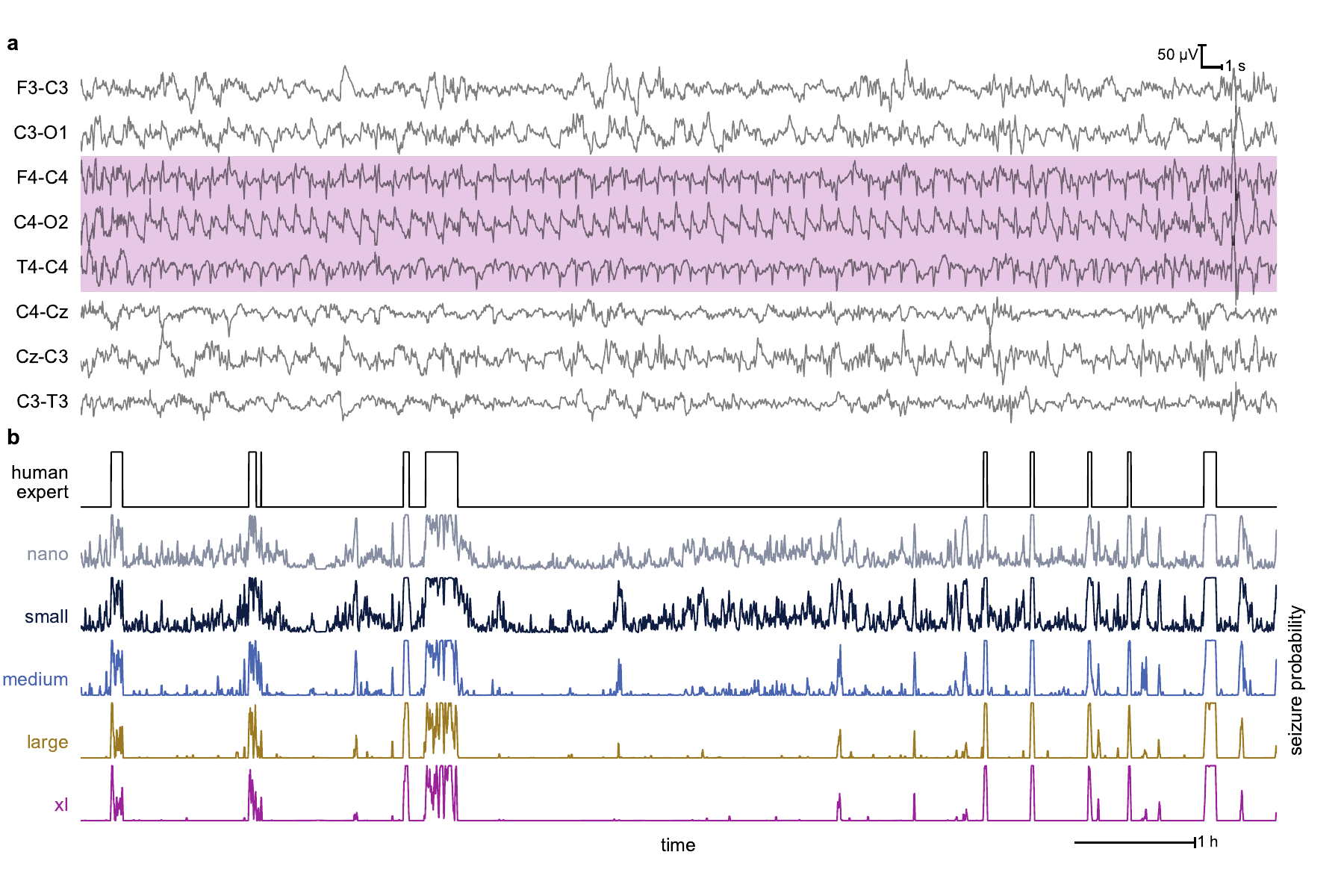}
    \caption{\small a) A 60 second sample of EEG with per channel seizure annotations shaded. We see that only 3/8 channels contain seizure. b) Annotation and model outputs for 10 hours from C4-O2 of the same EEG recording, from the development test set. The EEG sample in a) corresponds to the first 60s of the first seizure event in b). We see that for increasing model size, the models become more confident, suppressing the output for non-seizure periods while maintaining high agreement in seizure periods. This ease of interpretation would be beneficial to clinical implementation that use the real time model output trace.}
    \label{fig:eeg-prob}
\end{figure}

A total of 6,487 hours of multi-channel EEG was reviewed for seizure by two neonatal neurophysiologists. As each channel was reviewed and annotated separately, this equates to 50,299 hours of annotated EEG. Seizures were identified in 77 neonates with a total of 12,402 individual per-channel seizure events (see Fig.~\ref{fig:eeg-prob}a for example of per-channel annotations).

The per-channel annotations were used to develop a channel-independent algorithm. Different centres will use different protocols when recording EEG, ranging from a 1-channel amplitude-integrated EEG (aEEG) to a full 10:20 electrode array of 19 channels \cite{Stevenson2019}. Developing an AI model on a specific number of channels and a specific montage leads to models that are sensitive to that montage only. Electrodes may detach or become unusable due to artefact during recording. Sustaining a long-duration EEG recording, as is needed for seizure surveillance, without degradation of signal quality on some channels may be unrealistic, given the challenging recording environment of the NICU.

\begin{table}[h]
  \centering
    \footnotesize        
    \caption{\small Cohort demographics according to the EEG datasets. 
      % Cork datasets are internal and Helsinki refers to an open-access sets \cite{Stevenson2019}. 
      Data are presented as median (interquartile range) or number (percentage) unless otherwise
      stated. Number of neonates ($n$) is indicated when data not complete. The development set
      is used for training and testing the models and the validation sets are used for held-out
      testing. For the Helsinki dataset, clinical information is extracted from the associated metadata. 
      Common primary diagnoses of HIE, birth asphyxia, and stroke are included in the clinical demographics.}
  \begin{threeparttable}    
  \begin{tabular}{@{}llll@{}}
    \toprule
                                                    & Development Dataset     & \multicolumn{2}{c}{Validation Datasets}       \\
    \cmidrule(lr){2-2} \cmidrule(lr){3-4}
                                                    & Cork ($n=202$)          & Cork ($n=51$)       & Helsinki ($n=79$)       \\
    \midrule
    gestational age,                                & 40+0 (39+2 to 40+5)     & 40+4 (39+3 to 41+2) & 40 to 41\tnote{\S}      \\
    \makecell[c]{weeks+days}                        & \makecell[r]{[$n=197]$} &                     & \makecell[r]{[$n=78]$}  \\
    birth weight, kg                                & 3.50 (3.27 to 3.76)     & 3.50 (3.13 to 3.91) & 3.50 to 4.00 \tnote{\S} \\
                                                    & \makecell[r]{[$n=77]$}  &                     & \makecell[r]{[$n=64]$}  \\
    sex, female                                     & 86 (42.8\%)             & 22 (43.1\%)         & 35 (45.5\%)             \\
                                                    & \makecell[r]{$[n=201]$} &                     & \makecell[r]{[$n=77]$}  \\
    \multicolumn{2}{l}{\bf{Clinical demographics:}} &                         &                                               \\
    \midrule
    normal cohort\tnote{*}                          & 73 (36.2\%)             & 0 (0\%)             & 0 (0\%)                 \\
    hypothermia                                     & 28 (18.7\%)             & 13 (25.5\%)         &                         \\ 
                                                    & \makecell[r]{[$n=150]$} &                     &                         \\ 
    HIE,                                            & 63 (31.2\%)             & 27 (13.4\%)         & 29 (14.4\%)             \\
    \hspace{1em} mild/moderate/severe               & 23 / 26 / 14            & 8 / 13 / 6          & 3 / 8\tnote{\ddag} / 18 \\
    birth asphyxia                                  & 7 (3.5\%)               & 10 (5.0\%)          & 4 (2.0\%)               \\
    stroke                                          & 9 (4.5\%)               & 6 (3.0\%)           & 11 (5.4\%)              \\
    \multicolumn{2}{l}{\bf{EEG characteristics:}}   &                         &                                               \\
    \midrule    
    total duration, h                               & 6,487                   & 2,548               & 112                     \\
    duration\tnote{\dag} , h                        & 7.4 (1.1 to 51.6)       & 36.4 (21.4 to 74.4) & 1.24 (1.06 to 1.59)     \\
    start\tnote{\dag} , h                           & 13.8 (10.3 to 27.8)     & 4.6 (3.0 to 17.9)   & $<168$                  \\
    seizures\tnote{\dag}                            & 77 (38.1\%)             & 24 (47.1\%)         & 39 (49.4\%)             \\
    annotated seizure events                        & 12,402                  & 1,572\tnote{\P}     & 450\tnote{\P}           \\
    \bottomrule
  \end{tabular}
  \begin{tablenotes}
  \item {\footnotesize Key: HIE, hypoxic-ischaemic encephalopathy.}
        \item[*] {\footnotesize EEG collected from healthy neonates in the postnatal ward \cite{Korotchikova2016}.}
  \item[\dag] {\footnotesize per neonate; EEG start is time from birth to first EEG recording.}
  \item[\S] {\footnotesize mode, as data are categorical}
  \item[\ddag] {\footnotesize category includes moderate and mild/moderate as defined in the
      metadata \cite{Stevenson2019}
    \item[\P] average number for the 3 reviewers.}
  \end{tablenotes}
\end{threeparttable}
\label{tab:clin_demographics}
\end{table}

\subsubsection{Held-out EEG validation sets}

To validate the performance of our algorithms we tested on two held-out, unseen datasets. The first dataset is a cohort consisting of EEG from 51 term neonates with mixed aetiologies at risk of seizures \cite{Stevenson2015}. EEGs were reviewed independently by three international EEG experts, with a high level of agreement \cite{Stevenson2015}. Although the EEG data is collected in the same location as the development dataset (CUMH), there is no reviewer overlap between this and the development dataset. 

The second validation dataset is an open-access neonatal EEG dataset with seizure annotations \cite{Stevenson2019}. Again, this was reviewed by three EEG experts. The dataset consists of EEG from 79 term neonates with mixed aetiologies. 
For both validation datasets, seizure annotations were global, a single label used to indicate seizure in one or more channels. We refer to the datasets according to their geographic location: the Cork and Helsinki validation sets. Table~\ref{tab:clin_demographics} includes demographic information on both datasets.

\subsection{Seizure detection model}
We develop a modern convolutional neural network, based on the ConvNeXt architecture \cite{convext}, for our seizure detection model. The ConvNeXt architecture improved on previous convolutional models in several ways to achieve state-of-the-art in computer vision. These improvements include the use of depthwise convolutionals, stacked $1\times1$ layers, and a patched input stem to achieve a computational-efficient design. We adapt the architecture to the 1-D setting of EEG time series, modelling short 16~s single channel segments. 

In order to test our hypothesis of increasing model scale leading to improved performance we implement several variants of the model related by a simple width and depth scaling paramaterisation. The variants range from a $38.7$~k parameter Nano model up to a $20.6$~m parameter Extra Large (XL) model; see Table~\ref{tab:model_variants} (Appendix) for further details.

All models are trained to maximise classification performance of the 16~s segments. The hyperparameters and pre- and post-processing are the same for each model. Despite a 50:1 class imbalance due to rarity of seizure events and the large difference in model scale, our custom optimisation scheme is able to perform well in each case.  Full details on the development of the detection models is described in Appendix~\ref{sec:model_details}.

\subsection{Evaluating performance}

We conduct a comprehensive evaluation of the model using two complementary approaches: 1) performance metrics using human annotations as the gold standard and 2) human-expert equivalence testing.
First, we include a range of performance metrics to avoid reliance on a single metric. Because of the many limitations associated with the area-under the receiver-operating-characteristic curve (AUC) \cite{Lobo2007,saito2015precision,Chicco2023a}, we rely on more balanced measures such Pearson's correlation and Matthew's correlation coefficient (MCC) \cite{Chicco2020}. We also include clinically-relevant measures such as false detections per hour (FD/h) and total seizure burden. A complete list of metrics is presented in Table~\ref{tab:metric_definitions} (Appendix).

Second, we evaluate performance relative to inter-rater agreement using a test developed for neonatal EEG seizure detection \cite{Stevenson2019hybrid,Tapani2019,Tapani2022}. This approach estimates the impact on inter-rater agreement of replacing a human with the AI model, as measured by metrics from the $\kappa$ family. If $\Delta \kappa$, a difference in inter-rater agreement when replacing human annotations with AI predictions, is not significantly different to 0, then the model is considered to be operating at expert-level performance. For a more detailed description of this test procedure see Appendix~\ref{sec:kappa_test}. We also developed an open-source framework, written in Python code, to enable fair and reproducible performance analysis (available at \url{https://github.com/CergenX/SPEED}).

\section{Results}

\subsection{Data and model scaling}
\label{sec:model_scaling}

\begin{figure}
    \centering
    \includegraphics[width=1\linewidth]{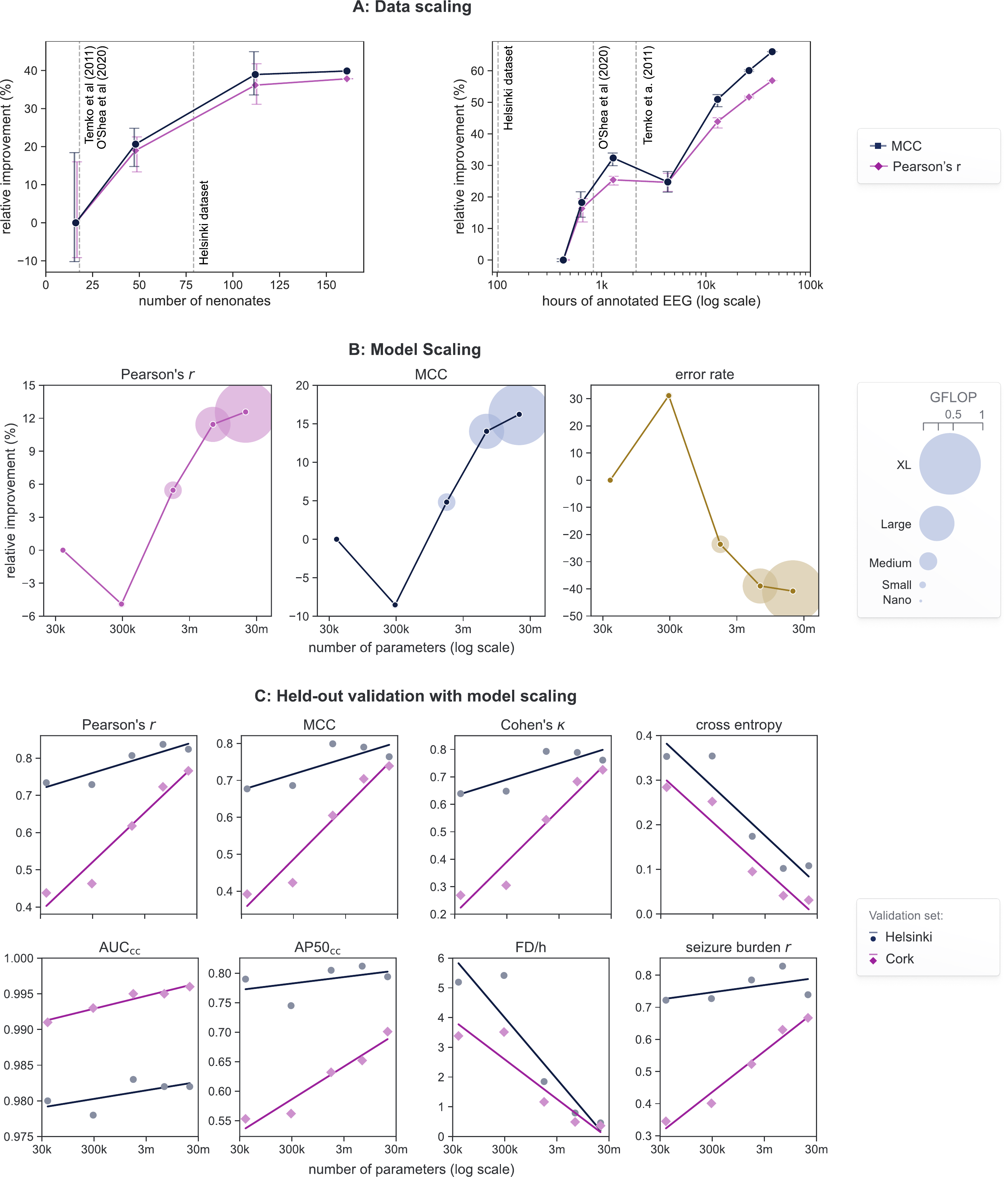}
      \caption{\small Scaling training data and model size yields approximate power-law performance gains.  Metrics calculated at the segment level for 20\% ($41/202$ neonates) of the development dataset in (a) and (b) and neonate level for held-out validation datasets in (c). \textbf{(a)}: Performance improves with increasing number of neonates and hours of annotated EEG (counted per channel) in training set; error bars denote min/max over 3 trials. Prominent datasets from the
  literature are included for comparison \cite{Stevenson2019,OShea2020,Temko2011}. \textbf{(b)}: Scaling model size over 3 orders of magnitude reveals typical deep-double descent
  pattern with a performance dip for the Small model before recovering for larger
  models. Marker size indicates computational cost in giga floating
  point operations (GFLOP).  \textbf{(c)}: Scaling model size on the held-out datasets from Cork and
  Helsinki. We include a linear fit to illustrate the predictability of performance increase.
  See \ref{sec:performance_metrics} for description of metrics used.}

    \label{fig:scaling_plots}
\end{figure}

We quantify the effect of increasing dataset size with sub-sampling by 1) EEG segment and 2) neonate. 
We do so by training a Medium model with 80\% of the development dataset and test on the left-out 20\%. We find that in both cases, there is significant performance gains, of up to 50\%, from scaling the data---illustrated in Fig.~\ref{fig:scaling_plots}a. Adding more EEG segments improves performance, even for datasets $>$20 times larger than the nearest published work, indicating that scaling data remains a powerful lever for improving models.

We evaluated a wide range of model scales, as described in Table \ref{tab:model_variants} (Appendix), from the 39~k parameter Nano variant up to the $>$500-times larger 21~m parameter XL model and find significant performance improvements. Fig.~\ref{fig:scaling_plots}b illustrates these improvement gains in MCC, correlation, and error rate suggesting that model scaling is indeed a viable path to better models for neonatal seizure detection. A representative sample of the model output in Fig.~\ref{fig:eeg-prob}b gives a qualitative sense of this performance improvement.

One notable feature is a large drop in performance for the first $\times10$ scaling from the Nano to Small model. This phenomenon has been observed repeatedly in other applications and is known as {\em deep double descent} \cite{nakkiran2019deep}. 
We also see some evidence of this in data scaling in Fig.~\ref{fig:scaling_plots}a from 1~k to 10~k hours of EEG. 

Fig.~\ref{fig:scaling_plots}c presents results for held-out validation sets across different model scales. As these datasets only have global annotations across all channels, we take the maximum over the per-channel outputs to produce a global prediction. This simplification may obscure some of the per-channel performance differences in the models. Nevertheless, the power-law trend of improvement with model scale is clear across many metrics, displaying a strong validation of the scaling hypothesis. We also find the appearance of the double-descent dip here again, although less pronounced, across several metrics on both datasets. 

We do, however, see indications of diminishing returns for some metrics for the Large and XL models on the Helsinki dataset. We investigate this further in Appendix~\ref{sec:distribution_shift}.

\subsection{Model Performance}
\label{sec:model_performance}
Table~\ref{tab:results_all} presents a comprehensive evaluation of the XL model across the 3 datasets.
In Table \ref{tab:helsinki-results} we also include the limited set of metrics available for direct comparison to the literature. Despite our relatively simplistic approach to translating from per-channel to global predictions, we still find that our models compare quite favourably to those published in the literature. This is true even for models that have been trained on the Helsinki dataset and report a cross-validation result. 

\begin{table}
    \centering
    \footnotesize
    \caption{\small Performance of the XL model on 3 datasets. Testing results are from 20\% of the
      development dataset. 
      Validation results are from held-out datasets from Cork and Helsinki (described in
      Table~\ref{tab:clin_demographics}.)
      Performance is assessed per-channel on the test dataset and globally (across all channels) on
      the validation datasets.
      All metrics are calculated by concatenating all EEG recordings. 
      Metrics for the held-out (validation) multi-annotator sets are based on unanimous consensus
      annotations. 
    }
    \begin{threeparttable}      
      \begin{tabular}{@{}llll@{}}
        \toprule
                                     & \thead{Test Set} & \multicolumn{2}{c}{\thead{Validation Sets}} \\
        \cmidrule(lr){2-2} \cmidrule(lr){3-4}
                                     & Cork ($n=41$)    & Cork ($n=51$)  & Helsinki ($n=79$)          \\
                                     & per-channel      & global channel & global channel             \\
        \midrule
        AUC                          & 0.978            & 0.996          & 0.982                      \\
        AP / AP50                    & 0.694 / 0.533    & 0.833 / 0.701  & 0.891 / 0.794              \\
        Pearson's $r$                & 0.723            & 0.766          & 0.824                      \\
        MCC                          & 0.648            & 0.739          & 0.764                      \\
        Cohen's $\kappa$             & 0.630            & 0.726          & 0.761                      \\
        Sensitivity/Specificity (\%) & 51.5 / 99.9      & 88.9 / 99.2    & 72.9 / 98.5                \\
        PPV/NPV (\%)                 & 82.0 / 99.6      & 62.0 / 99.8    & 84.9 / 96.8                \\
        FD/h                         & 0.053            & 0.363          & 0.459                      \\
        Seizure Burden, $r$          & 0.902            & 0.667          & 0.739                      \\
        \bottomrule
      \end{tabular}
      \begin{tablenotes}
        \footnotesize
      \item Key: AUC, area under the receiver operator characteristic curve; AP, average
        precision; AP50, average precision with recall $>$50\%; MCC, Matthew's correlation
        coefficient; PPV, positive predictive value; NPV, negative predictive value; FD/h, false
        detections per hour; $r$ represents correlation; $\kappa$ represents kappa. 
       \end{tablenotes}
     \end{threeparttable}
      \label{tab:results_all}     
\end{table}

\begin{table}
    \centering
    \footnotesize

    \caption{\small Comparison of proposed model and other published models tested on the Helsinki
      dataset. Our extra-large (XL) model achieves a new state-of-the-art on most metrics, even outperforming
      models trained directly on the Helsinki data. 
      In keeping with published methods, all but the concatenated AUC (AUC$_{\textrm{cc}}$) is
      evaluated on the subset for neonates ($n=39$) with seizures. 
      A more complete set of metrics on all 79 neonates is shown in Table~\ref{tab:results_all}. 
    }
    \begin{threeparttable}    
      \begin{tabular}{lccccc}
        \toprule
                                                    & all data ($n=79$) & \multicolumn{3}{c}{seizure only ($n=39$)}                                  \\
        \cmidrule(lr){2-2} \cmidrule(lr){3-5}
                                                    & AUC$_{\textrm{cc}}$        & AUC (median [IQR])                    & Cohen's $\kappa$ & FD/h            \\
        \midrule
        XL ConvNeXt (ours)                          & $\textbf{0.982}$  & $\textbf{0.996 (0.975$ - $1.000)}$ & $0.800$          & $\textbf{0.34}$ \\
        FcCNN \cite{OShea2020}                      & $0.956$           & -                                     & -                & -               \\
        ResNet \cite{Daly2021}                      & $0.964$           & -                                     & -                & -               \\ 
        SVM--Cork  \cite{Temko2011}                 & -                 & $0.961\ (0.869 - 0.990)$              & -                & $1.00$          \\
        SVM--Helsinki \cite{Tapani2019}\tnote{\dag} & $0.955$           & $0.988\ (0.931 - 0.998)$              & -                & $0.86$          \\
        GAT \cite{Raeisi2023}\tnote{\dag}           & -                 & $0.993\ (0.964 - 0.995)$              & $\textbf{0.880}$ & $0.86$          \\
        \bottomrule
    \end{tabular}
    \begin{tablenotes}
      \footnotesize
      \item[\dag] Leave-one-out (LOO) testing result using the Helsinki dataset. 
      \item Key: IQR, interquartile range; AUC, area under the receiver operator characteristic curve; FcCNN, fully
        convolutional neural network; SVM, support vector machine; GAT, graph attention network;
        FD/h, false detections per hour. 
      \end{tablenotes}
    \end{threeparttable}      
    \label{tab:helsinki-results}
\end{table}

\subsection{Expert-level agreement}

The XL model attains expert-level equivalence on both Cork and Helsinki validation datasets.
In both cases, the change in agreement by replacing a human expert with the AI model predictions was consistent with 0: $\Delta \kappa = -0.094$ (95\% CI: -0.189, 0.005) for Cork and $\Delta \kappa = -0.082$ (-0.156, 0.002) for Helsinki. For the Helsinki dataset, the Medium model also reaches this benchmark and the Large model is just narrowly rejected but neither model achieves this benchmark on the Cork validation dataset. Results for all models are presented in Table \ref{tab:bootstrap_tests} (Appendix).

\section{Discussion}
We have developed a state-of-the-art convolutional neural network for neonatal seizure detection,
improving substantially upon previous published results. We also have verified our hypothesis that scaling is a hitherto under-utilised lever for
performance improvement in neonatal EEG analysis. Scaling both the dataset size, by neonate and by
duration of EEG, yielded up to 50\% increases in MCC. Scaling model
sizes similarly delivered significant performance improvements of up to 15\% in MCC. The result of these improvements is that our best model, the 21~m parameter XL variant
of the ConvNeXt architecture, attains expert-level equivalence with the EEG experts on two independent, fully held-out validation sets ($\Delta \kappa \neq 0$ rejected with $p>0.05$).

Much of the literature focuses on methodological improvements, with specialised architectures
trained on very small datasets yielding incremental gains
\cite{Caliskan2021,Tanveer2021,Gramacki2022,Raeisi2022,Raeisi2023}. Our work challenges this
approach and suggests a more promising path to expert-level models is through data and model scale. A key part of the model scaling strategy is
designing an architecture with computational efficient scaling. Failure to do so can lead to prohibitively expensive training iterations. Scaling the fully-convolutional neural
network model \cite{OShea2020}, for example, to an equivalent size of the XL model would require more than $6$ times the computational
load. 

Our scaling results also challenge the conventional wisdom that increasing model size will eventually lead to overfitting and decreased generalisation performance. Indeed to date, most research in neonatal EEG has focused on relatively small models, with less than 100~k parameters \cite{Stevenson2019hybrid,OShea2020,Daly2023}. Despite this, model scaling well past the point of over-parameterisation has been a key feature of recent AI progress, for example see \cite{kaplan2020scaling, hoffmann2022training, nakkiran2019deep}. This observation that performance will initially decline before improving with scaling is known as deep-double descent and was found to occur across a
range of tasks, model architectures, and optimisation methods \cite{nakkiran2019deep}.  
Fig.~\ref{fig:scaling_plots}b illustrates this finding in all metrics with a decrease in performance for the Small model comparative to the smaller Nano model.
We also see indications of this in data scaling (Fig.~\ref{fig:scaling_plots}a), where increasing the
size of the training dataset actually decreases performance before improving again with more data. This surprising finding is a corollary of the deep-double descent effect on model scale and was also observed in \cite{nakkiran2019deep}. If operating in a narrow scale range, on the left-hand side of the double-descent dip, it is
understandable that smaller models and datasets would seem optimal (as found in \cite{OShea2020}). 
However, an exploration of a much larger scale range, as we show here, yields substantial benefits
by moving past the double-descent trap. 

A limitation in the neonatal
seizure detection literature is that AUC is almost always presented as the lead---and often only---performance metric
\cite{Temko2011,Ansari2019,pmlr-v126-isaev20a,OShea2020,Daly2021,Caliskan2021,Tanveer2021,Gramacki2022,Raeisi2022,Daly2023,Raeisi2023}.
This metric can be misleading for many reasons \cite{Lobo2007,saito2015precision,Chicco2023a}.
For example, with large class imbalance, as is the case for electrographic seizures, false positives are obscured. To illustrate this, our worst performing model (Nano) has an AUC of 0.980 on the Helsinki set, exceeding the best reported value of
$0.964$ \cite{Daly2021}. Our XL model improves on this only slightly to $0.982$ but has
approximately 10-times fewer FD/h and achieves expert-level agreement on both
held out datasets. The Nano model, in contrast, is far from achieving expert-level agreement:
$\Delta \kappa$ is approximately 5-times (2-times) larger on Cork (Helsinki) validation datasets.

Addressing this limitation, we present a comprehensive set of metrics for continuous and binary variables, including more balanced measures of performance, such as MCC, Pearson's $r$, and Cohen's $\kappa$
\cite{Lobo2007,saito2015precision,Chicco2020,Chicco2023a}, in addition to metrics with more
clinical relevance, such as FD/h, correlation with seizure burden, and expert-level equivalence testing. We have developed an open-source framework for metric calculation to assist with transparency in reporting of performance for this field. 

We have also highlighted the utility of developing models with per-channel annotations, making the
algorithm adaptable to different clinical montage requirements or protocols. 
Fig.~\ref{fig:sz_perchannel_stats} (Appendix) illustrates the heterogeneous time-varying nature of seizure
focus among EEG channels. As a result, global labels will obfuscate important channel differences,
similar to injecting noise into the training data. Although global labels, or weak labels
\cite{OShea2020}, are easier to annotate, they present only summary information without detail and
therefore fail to maximise the full potential of the valuable EEG data. 
By providing a strong training label, Fig.~\ref{fig:montage-degradation} (Appendix) shows that per-channel
models are flexible to different montages and even robust to large amounts of data loss, as is
likely to occur in a clinical environment. 

We found evidence of a distribution shift on the Helsinki validation set.
Returns on model scaling appears to diminish after the Medium model with the best model becoming
metric dependent, indicating that the gains for the Large and XL models don't transfer as well to
this dataset (see Fig.~\ref{fig:scaling_plots}c). Analysis in Appendix \ref{sec:distribution_shift} indicates that the Large and XL models are learning
something specific to the early-EEG group (postnatal age $<$ 1 week) compared to the late-EEG group
($>$ 1 week). We speculate that this could be related to subtle differences in the EEG
waveforms associated with either postnatal age or, more likely, with primary diagnosis such as HIE
or stroke versus other primary diagnoses such as sepsis, meningitis, or recovery post cardiac
surgery \cite{Stevenson2019}. This suggests that future development of seizure detectors could
benefit from more diverse training data, recorded from neonates at different postnatal ages and with
more varied pathologies and seizure aetiologies other than HIE and stroke. 

The key result of this work is---for the first time---a thorough demonstration of an expert-level
neonatal EEG seizure detector. Although this claim has been made before \cite{Stevenson2019hybrid} it was accompanied by some important caveats. First, it was a
cross-validation result and not a held-out dataset. Second, this model failed to reach expert-level equivalence when validated on a held-out set
\cite{Tapani2022}.
Third, statistical equivalence was found for only one $\Delta \kappa_a$, when replacing one
expert, and not for the overall $\Delta \kappa$, an average over the 3 annotators, as our test
finds. In our work, in contrast, we report statistical equivalence to experts on two different fully
held-out datasets with a combined number of 130 neonates with over 2.7~k hours of EEG.
For these reasons, we believe that our claim of expert-level equivalence is the first of its
kind for neonatal seizure detection.

This study is not without limitations. The observed distribution shift on the Helsinki validation set
suggests the XL model works best within the first week of life. Although seizures are most common
during this period \cite{Tekgul2006,Soul2018,Pisani2021}, we should not assume that this covers all
possible use cases. Another possible limitation is that our development dataset is from one centre. A promising direction for improvement on both counts is to train on a more diverse multi-centre dataset of EEG with recordings from a larger postnatal time range. And lastly, although we show that the proposed model attains expert-level
agreement on our retrospective validation sets, a clinical investigation of the algorithm cotside is
the best way to evaluate utility.

In conclusion, we find strong evidence that scaling training data and model size improves
performance for neonatal EEG seizure detection.
Held-out validation, on datasets with a combined total 2.7~k hours of
multi-channel EEG from 130 neonates, found accurate and reliable generalisation performance. Achieving expert-level
performance demonstrates readiness for clinical validation. Automated analysis of long-duration EEG
facilitates increased seizure surveillance for at-risk neonates. This, in turn, can assist in timely
neuroprotective strategies to help improve long-term outcomes for vulnerable neonates in critical
care.

\section*{Funding:} GB was supported by a Wellcome Trust Innovator Award (209325/Z/17/Z).

\renewcommand{\appendixname}{Appendix}
\appendix

\renewcommand{\appendixname}{Appendix}
\renewcommand{\appendixpagename}{{\Large Appendix}}
\begin{appendices}
\appendixpage

In this Appendix, we add more details on the developmental dataset used for training (Section \ref{supp:dev_dataset}) and detail design and development of the convolutional neural network (Section~\ref{sec:model_details}). We also provide a detailed description of the performance metrics used to assess the models (Section~\ref{supp:perf_analysis}). Lastly, Section~\ref{supp:add_analysis} provides additional analysis of model performance related to expert-level tests, duration of seizure events, a distribution shift for the Helsinki validation set, and stress-testing of the method to montage degradation.

\section{EEG development dataset}
\label{supp:dev_dataset}

% Median (IQR) = 1.28 (1.02, 6.17) hours; range = 0.31 - 12.28 hours
EEG records were obtained from {\em Neobase}, a fully anonymised database of neonatal EEG recording obtained from the Cork University Maternity Hospital, Ireland. Informed consent was obtained from the parents or guardians  for all research studies and ethical approval was obtained from the Clinical Research Ethics Committee of the Cork Teaching Hospitals. One of the following EEG machines was used: the Neurofax EEG-1200 (Nihon Kohden), NicoletOne ICU Monitor (Natus, USA), or the Lifelines EEG (iEEG Lifelines, Stockbridge, United Kingdom). Sampling frequencies were set at 200 Hz, 256 Hz, or 500 Hz depending on the machine. EEG signals were recorded from the frontal (F3/F4, Fp1/Fp2, or Fp3/Fp4), temporal (T3/T4), central (C3/C4 and Cz), and occipital (O1/O2) or parietal (P3/P4) regions. 

EEG recording commenced as soon as possible after birth and continued for hours or days. EEGs were recorded from term neonates with mixed aetiologies at risk of seizures in the neonatal intensive care unit (NICU) in most cases. We also include a control subset of healthy term newborns recorded in the postnatal wards ($\leq 2$~hours) which was used as part of training data \cite{Korotchikova2009,Korotchikova2016,Raurale2020,Garvey2021}.

The following bipolar montage was used for reviewing and annotating the EEG for seizures: F4--C4, C4--O2, F3--C3, C3--O1, T4--C4, C4--Cz, Cz--C3, and C3--T3. For the control cohort of healthy newborns, without left and right central electrodes, the montage was set to F4--T4, Cz--O2, F3--T3, Cz--O1, T4--Cz, Cz--T3.
Total seizure burden ranged from 0.79 to 981.98 minutes, with a median (IQR) of 50.91 (20.31 to 113.08) minutes. A total of 12,402 per-channel seizure events were annotated, with a median (IQR) of 48 (19 to 144) distinct seizures events per neonate. Demographic and clinical data are presented in Table~\ref{tab:clin_demographics}.

To estimate inter-rater agreement, two neonatal neurophysiologists reviewed a subset of the dataset, consisting of EEG from 13 neonates. Cohen's $\kappa$ indicated high inter-rater agreement, with a median $\kappa$ of 0.808 (IQR: 0.702 to 0.874; range: 0.548 to 0.990). Although this is calculated on a per-channel rather than global annotation, agreement is in keeping with the previously reported estimates of inter-rater agreement: $\kappa=0.767$ for the Helsinki dataset \cite{Stevenson2019} and $\kappa=0.827$ for a Cork/London dataset \cite{Stevenson2015}; both assessments used Fleiss $\kappa$ to account for the 3 reviewers. 

\begin{figure}[ht!]
  \centering
  \includegraphics[width=1\textwidth]{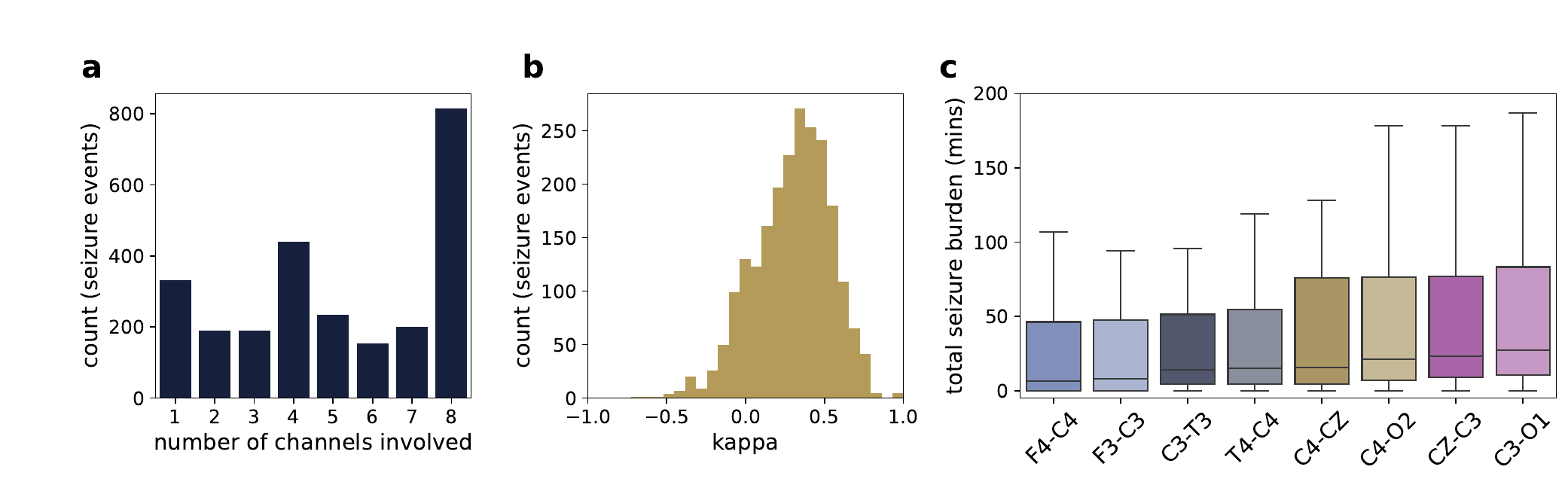}
  \caption{\small Summary of per-channel EEG seizure annotations for 77 neonates. 
    (a): number of channels involved in each seizure event. 
    (b): agreement among seizure annotations across channels for each seizure event,
    as quantified by Fleiss $\kappa$. 
    (c): total seizure duration for each neonates' EEG estimated from each channel separately. For
  a small number of EEGs, F3 is replaced by Fp1 or Fp3; and likewise, F4 is replaced by Fp2 or Fp4.} 
  \label{fig:sz_perchannel_stats}
\end{figure}

Analysis of the per-channel annotations indicate a high degree of variability in the number of EEG channels involved in each seizure event and in the variability of the time-synchronization of seizures across channels, as illustrated in Fig.~\ref{fig:sz_perchannel_stats}.
This figure also indicates that seizure burden is roughly independent of EEG channel, although the frontal channels (F3--C3 and F4--C4) appear to have a slightly lower burden compared to the other channels.

\section{Model development}
\label{sec:model_details}
In this section we describe our convolutional neural network based seizure-detection algorithm, how it was trained, and the pre- and post-processing procedures.

\subsection{ConvNeXt architecture}

\begin{figure}
    \centering
    \includegraphics[width=0.3\linewidth]{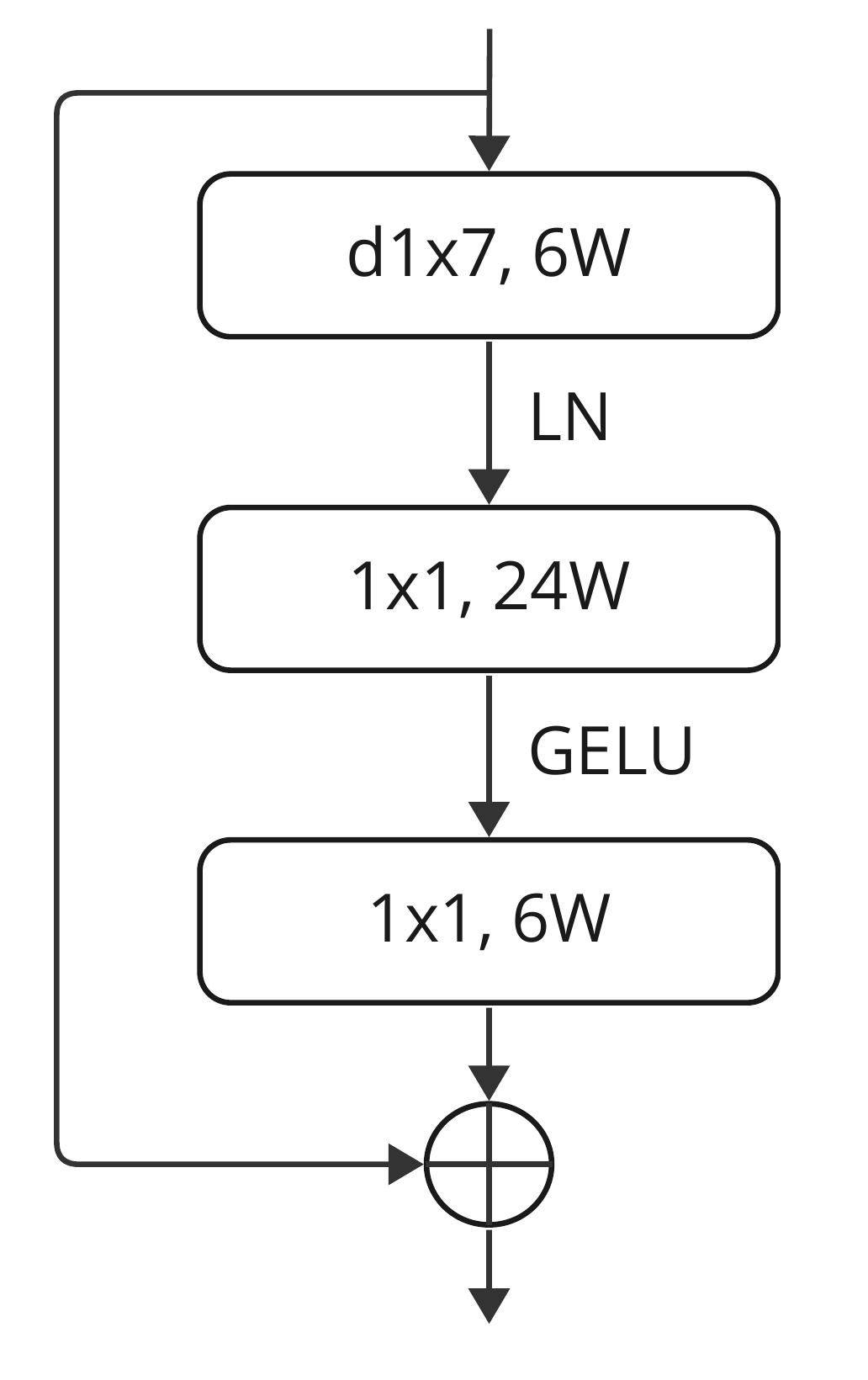}
    \caption{\small ConvNeXt block, adapted from \cite{convext}. Here the $W$ is
      an integer parameter we use to control the width of our models. The block includes three
      convolutional layers, one with downsampling (indicated by the $d$) and one-dimensional kernel
      of length 7 samples, followed by two $1\times 1$ convolutional layers, an equivalent implementation of a multi-layer
      perceptron. Notable features of the use of layer normalisation (LN) rather than batch normalisation and a Gaussian
      error linear unit (GELU) instead of the rectified linear unit (ReLU).}
    \label{fig:block}
  \end{figure}

We adapt the ConvNeXt architecture \cite{convext}, originally designed for 2D computer vision applications, to our 1D time-series EEG data. This architecture was systematically designed for efficiency and performance. Taking inspiration from the recent success of vision transformer architectures it was designed with purely convolutional components and achieved state-of-the-art performance across several computer-vision tasks \cite{convext}. The basic building block of the model is shown in Fig.~\ref{fig:block}. Notably, the use of depth-wise convolution and stacked $1 \times 1$ convolutional layers contribute to increased computational efficiency without sacrificing accuracy.

  The detailed architecture is described in Table \ref{tab:arch}. Our parameterisation defines the network by 2 parameters: $D$ for depth and $W$ for width. Due to the residual structure, simply varying these two integer values allows for easy creation of model variants at different scales without any further adjustments. In this work, we explore models ranging in scales from 38.7~k to 20.6~m parameters; see Table~\ref{tab:model_variants} for the depth--width parameter settings for each model. 

\begin{table}
    \centering
    \footnotesize
    \caption{\small Model architecture for the proposed ConvNeXt model. Parameters ($D$, $W$) define the
      depth ($D$) and width ($W$) of the network. The network starts with the stem, followed by multiple
      stages, and finishes with average pooling and a linear layer to combine features. The input
      tensor is an array of 1,024 samples for 16 s of EEG sampled at 64 Hz.} 
    \begin{threeparttable}    
    \begin{tabular}{lccc}
      \toprule
          &        & input tensor    & output tensor   \\
      component    & description          & (dimension)         & (dimension)         \\
      \midrule
      stem         & $1\times4$, stride 4 & $1\times 1\times 1024$  & $6W\times 1\times 256$  \\
      stage 1      & \cnxblock{6}{24}{}   & $6W\times 1\times 256$  & $6W\times 1\times 256$  \\
      stage 2      & \cnxblock{12}{48}{}  & $6W\times 1\times 256$  & $12W\times 1\times 128$ \\
      stage 3      & \cnxblock{24}{96}{3} & $12W\times 1\times 128$ & $24W\times 1\times 64$  \\
      stage 4      & \cnxblock{48}{192}{} & $24W\times 1\times 64$  & $48W\times 1\times 32$  \\
      average pool  & $1\times32$          & $48W\times 1\times 32$  & $48W\times 1\times 1$   \\
      linear layer & $48W \times1$        & $48W\times 1\times 1$   & $1$ \\
      \bottomrule
    \end{tabular}
  \end{threeparttable}
    \label{tab:arch}  
\end{table}

\begin{table}
    \centering
    \small
    \caption{\small Model variants explored in this work. The model scale covers almost 3 orders of magnitude in parameter count.}    
    \begin{threeparttable}
      \begin{tabular}{lccrr}
        \toprule
        model       & depth & width & parameters & computation \\
                    & ($D$) & ($W$) & (count)    & (FLOP)      \\
        \midrule
        Nano        & 1     & 1     & 38.7 k     & 1.9 m       \\
        Small       & 2     & 2     & 289.2 k    & 14.4 m      \\
        Medium      & 3     & 4     & 1.7 m      & 84.3 m      \\
        Large       & 3     & 8     & 6.7 m      & 335.4 m     \\
        Extra Large & 6     & 10    & 20.6 m     & 1 G         \\
        \bottomrule
    \end{tabular}
    \begin{tablenotes}
      \footnotesize
      \item Key: FLOP, floating point operations
    \end{tablenotes}
  \end{threeparttable}
  \label{tab:model_variants}
\end{table}

\subsection{Training methods}

\paragraph{Addressing Imbalance}
For long-duration continuous recordings, seizure events typically occupy a small fraction of recording time, with the majority of the EEG being seizure free. A study by Rennie and colleagues described a median (IQR) total seizure burden of 69 (28 to 118) minutes over a median (IQR) of 70 (31 to 97) hours of EEG recording \cite{Rennie2018}. Additionally, not all neonates with EEG monitoring will have seizures: the same study found that 139 from 214 neonates did not have recorded electrographic seizures, despite the long duration of monitoring. Our development dataset reflects this imbalance, with an approximate class imbalance of 50:1. This imbalance can present a challenge for training machine-learning models, as the models can become biased towards the majority class.

The 3 most common ways to deal with this are oversampling the minority class, undersampling the majority class, and re-weighting the loss function. Oversampling is computational demanding, and for large datasets such as long-duration EEG recordings, unappealing and wasteful of expensive computational resources. Undersampling is also wasteful, as a large proportion of the diverse EEG records are discarded. Loss re-weighting is usually a good option but in our case such a large imbalance can result in large loss values which, even with gradient clipping, can de-stabilise the learning.

Instead, we develop a new approach to this problem: we keep all data and dynamically undersample the non-seizure examples at random during training. From one training epoch to the next the model will see a different sample of the non-seizure data but the same seizure data. By selecting a different random sample of non-seizure data per training epoch, all of the non-seizure data will be exposed during training with a sufficient number of training epochs. In practice, we found that dynamically undersampling at a ratio of 5:1 and combining loss re-weighting to account for this imbalance was the most stable and efficient implementation. 

\paragraph{Learning optimisation}

All models are trained with the same hyperparameters using AdamW on a learning-rate schedule. The learning-rate schedule follows a variant of the 1-cycle policy \cite{Smith2019} with 4 phases: warmup, freeze-at-max, cooldown, then freeze-at-min. The learning rate changes logarithmically during the warmup and cooldown phases. In our experiments we found this schedule reliably led to training convergence. This eliminated the need for early-stopping based on monitoring of validation loss. Although common practice in many machine-learning applications, we have found this to be unreliable. The large variability among neonates resulted in the early stopping condition being highly sensitive to the choice of babies in the validation set. An approach to mitigate this is to use more than one $k$-fold \cite{OShea2020}, but this results in several models that need to be ensembled somehow. Problematically, this sensitivity of the model to the validation set raises questions about generalisation to unseen data when using this method. Additionally, a consequence of the deep-double descent phenomenon which we observed in Sec \ref{sec:model_scaling} is that early stopping will only select the best model in the special case of the model size and dataset size are critically balanced \cite{nakkiran2019deep}.

\paragraph{Data augmentation}

To improve model robustness we developed and experimented with several data augmentation techniques. This consisted of several signal processing transformations: magnitude scaling, magnitude warping, jitter, time warping, and spectral-phase randomisation. In addition, generic transformations such as flip, cutmix \cite{yun2019cutmix}, cutout \cite{devries2017cutout}, and mixup \cite{zhang2018mixup} were applied. The parameters of each augmentation were manually adjusted to ensure all transformations were label preserving. Different probabilities were assigned to each transformation for a given batch. From our experimentation, only flip and cutout gave consistent improvements in performance and were therefore included in the model development presented here.

\subsection{Pre- and Post-Processing}

Pre-processing of the EEG included the following stages: band-pass filtering within the 0.3--30 Hz passband, downsampling to 64~Hz, and removal of some artefacts. These artefacts were either periods of contiguous zeros, caused by checking the impedance of electrode scalp contact, or periods of excessive high-amplitude activity, defined by a standard deviation greater than 1~mV for each segment. EEG was divided into 16~s segments with a step size of 4~s. These segments were used for training with a label of seizure if $\ge8$~s of the segment was annotated as seizure and non-seizure otherwise.

When testing the model with a full EEG recording, we process 16~s segments with a step size of 0.25~s. The continuous-valued output of the model is then smoothed with a 32~s rectangular window. From this probability-like output, we apply the standard threshold of 0.5 to generate the binary decision mask. We deliberately restrict our post-processing to be simple and limited in contrast to some more involved schemes in previous work \cite{Temko2011, OShea2020, Tapani2019}. While approaches like adding a collar to detected events or optimising the threshold can help with some metrics on some datasets \cite{Temko2011,Tapani2019,OShea2020}, we believe the best way to generalise well to other datasets is to rely on the model to learn the start and end of seizure events directly from the data.

All models were designed and built using the development dataset. The set was divided with a random 80:20 split of neonates: 80\% of neonates' EEG used for training and 20\% for testing. When development was finalised the models were then trained on all the development dataset and tested on the held-out validation sets. There was no back-and-forth between model development and testing on the held-out datasets.  

\section{Performance analysis}
\label{supp:perf_analysis}
\subsection{Metrics}
\label{sec:performance_metrics}

To assess and compare performance of different models across different datasets we use a range of metrics. AUC is the most commonly used metric in this field, yet is plagued with shortcomings including sensitivity to class imbalance \cite{Lobo2007,saito2015precision,Chicco2023a}. We opt to include more transparent measures, such as Pearson's correlation coefficient $r$ and its binary equivalent, MCC \cite{Chicco2020,Chicco2023a}. We also include the area under the precision--recall curve, a metric known to be relevant for rare-event detection tasks \cite{saito2015precision}. And we include a version which excludes the region for which recall $<50\%$, as this lower region is not useful for clinical application and can also be sensitive to outliers. 

A complete breakdown of the metrics used in this work are shown in Table \ref{tab:metric_definitions}. With multiple annotators, as we have for both our validation sets, we follow the convention of using a consensus annotation \cite{Tapani2019,OShea2020,Raeisi2023}.

To enable reproducible research, we provide computer code to generate the performance metrics used in the study. This evaluation framework generates a suite of metrics for any given prediction and ground-truth labels and has a default assessment specifically for the open-access Helsinki validation dataset \cite{Stevenson2019}. This framework, written in Python and known as SPEED (Seizure Prediction Evaluation for EEG-based Detectors), is freely available at \url{https://github.com/CergenX/SPEED}. 

\begin{table}
    \centering
    \small
    \caption{\small Description of metrics used in this work. Note whenever $cc$ subscript is used it means the value was computed by concatenating all recordings.}
    \begin{threeparttable}
      \begin{tabular}{lll}
        \toprule
        variable type     & name & description \\
        \midrule
        \multirow{5}{*}{continuous} 
        & AP               & average precision (area-under the precision--recall curve)  \\
        & AP50             & AP for recall $>0.5$ $\times 2$ to normalise to range (0,1)  \\
        & Pearson's $r$   & Pearson's correlation coefficient for ($p$, $y$) \\
        & cross entropy    & $\textrm{mean}(y \log(p) +(1-y) \log (1-p))$ \\
        & AUC              & area under the receiver-operator-characteristic curve    \\
        \midrule
        \multirow{10}{*}{binary} 
        & PPV              & positive predictive values (precision); $TP/(TP+FP)$ \\
        & NPV              & negative predictive values; $TN/(TN+FN)$ \\
        & sensitivity      & $TP/(TP+FN)$ (recall) \\
        & specificity      & $TN/(TN+FP)$ \\
        & error rate       & $(FP+FN)/N$; $1-\textrm{accuracy}$ \\
        & MCC              & Matthews correlation coefficient  \\
        & Cohen's $\kappa$ & measure of pairwise agreement accounting for chance \\
        & Fleiss' $\kappa$ & generalisation of Cohen's $\kappa$ to $>$2 annotators \\
        & FD/h             & false event detection per hour \\ %(no overlap with true seizure)\\
        & seizure burden, $r$ & correlation coefficient for hourly estimate of predicted \\
        &                    & versus true seizure burden in mins/h \\                
        \bottomrule
      \end{tabular}
      \begin{tablenotes}
        \footnotesize
        \item Key: $y$, true label; $p$, model prediction probability; TP, true positives; FP, false positives; TN, true negatives; FN, false negatives; N, total predictions 
      \end{tablenotes}
    \end{threeparttable}
    \label{tab:metric_definitions}
\end{table}

\subsection{Expert-level test}
\label{sec:kappa_test}
The metrics presented in Table~\ref{tab:metric_definitions} are useful for assessing model performance compared to a gold-standard annotation (single expert) or a consensus annotation (multiple experts). 
However, these do  not capture the disagreement among experts or evaluate performance in relation to the level of agreement among the experts. While the metrics discussed here are useful for research purposes in evaluating and comparing models, an assessment in relation to inter-rater agreement is more relevant for clinical adoption. 

We use an existing method proposed for this application, evaluation of models in relation to inter-rater agreement for neonatal EEG seizure detection \cite{Stevenson2019hybrid,Tapani2019,Tapani2022}. The method measures the impact of replacing each expert with the AI-model predictions and quantifying the difference in inter-rater agreement. This approach uses metrics from the $\kappa$ family to account for agreements by random chance, and estimates the impact of replacing a human with the model. We define this difference in agreement for our 3 annotator held-out datasets as
\begin{equation*}
\centering
  \Delta \kappa_l = \kappa_{\textrm{experts}} - \kappa_{\textrm{AI},a} \quad \textrm{for $a=1,2,3$}
\end{equation*}
where $\kappa_{\textrm{experts}}$ is the inter-rater agreement among the 3 experts and $\kappa_{\textrm{AI},a}$ is the agreement with 2 experts and the AI predictions for the 3 possible combinations. All $\kappa$ values are estimated using Fleiss $\kappa$. An overall difference in agreement, $\Delta \kappa$, is estimated as the mean value of $\Delta \kappa_a$ over the 3 experts. The condition of $\Delta \kappa = 0$ indicates that the AI predictions do not change inter-rater agreement and therefore can be considered equivalent \cite{Tapani2019,Tapani2022}. To test whether $\Delta \kappa = 0$, we follow the process of generating a distribution of $\Delta \kappa$ by bootstrapping with 1,000 iterations randomly sampling neonates' EEG. From this distribution, if the 95\% confidence interval (CI) includes 0 then we accept the null hypothesis that the model predictions do not significantly alter inter-rater agreement. Adherence to this condition establishes expert-level performance for the AI model. 

\section{Additional analysis}
\label{supp:add_analysis}
\subsection{Expert-level test results}
Table~\ref{tab:bootstrap_tests} presents the expert-level tests for all 5 models. The target of expert-level equivalence is only achieved on both validation sets with the XL model. For the smaller models, such as the Nano and Small, this benchmark is well out of reach ($p<0.001$).

\begin{table}[ht!]
    \centering
    \footnotesize
    \caption{\small Estimates of the $\Delta \kappa$, the change in level of agreement by replacing
      a human expert with the AI model predictions.
      Results are described for both Cork and Helsinki held-out datasets for all model sizes.      
      Distributions are estimated from $1,000$ bootstrap samples and confidence intervals (CI)
      including zero and a $p$-value $>$0.05 indicate no difference in inter-rater agreement. The
      extra-large (XL) model passes the test for expert equivalence on both datasets.}
    \begin{threeparttable}
      \begin{tabular}{lrrrr}
        \toprule
               & \multicolumn{2}{c}{Cork ($n=51$, 2.5k hours)} & \multicolumn{2}{c}{Helsinki ($n=79$, 102 hours)}       \\
         \cmidrule(lr){2-3} \cmidrule(lr){4-5}
        model  & $\Delta \kappa$ (95\% CI)     & $p$-value & $\Delta \kappa$ (95\% CI)      & $p$-value \\
        \midrule
        Nano   & -0.424 (-0.572 to -0.275)     & $<0.001$  & -0.132  (-0.206 to -0.060)     & $<0.000$  \\
        Small  & -0.387 (-0.521 to -0.252)     & $<0.001$  & -0.126 (-0.193 to -0.058)      & $<0.001$  \\
        Medium & -0.195 (-0.324 to -0.066)     & $0.003$   & \bf{-0.052  (-0.118 to 0.010)} & $0.099$   \\
        Large  & -0.114 (-0.221 to -0.008)     & $0.035$   & -0.066 (-0.126 to -0.001)      & $0.047$   \\
        XL     & \bf{-0.094 (-0.189 to 0.005)} & $0.063$   & \bf{-0.082 (-0.156 to 0.002)}  & $0.055$   \\
      \bottomrule
    \end{tabular}
    \end{threeparttable}
    \label{tab:bootstrap_tests}
  \end{table}

\subsection{Event duration analysis}

\begin{figure}[h]
    \centering
    \begin{subfigure}[b]{0.45\textwidth} 
        \includegraphics[width=\textwidth]{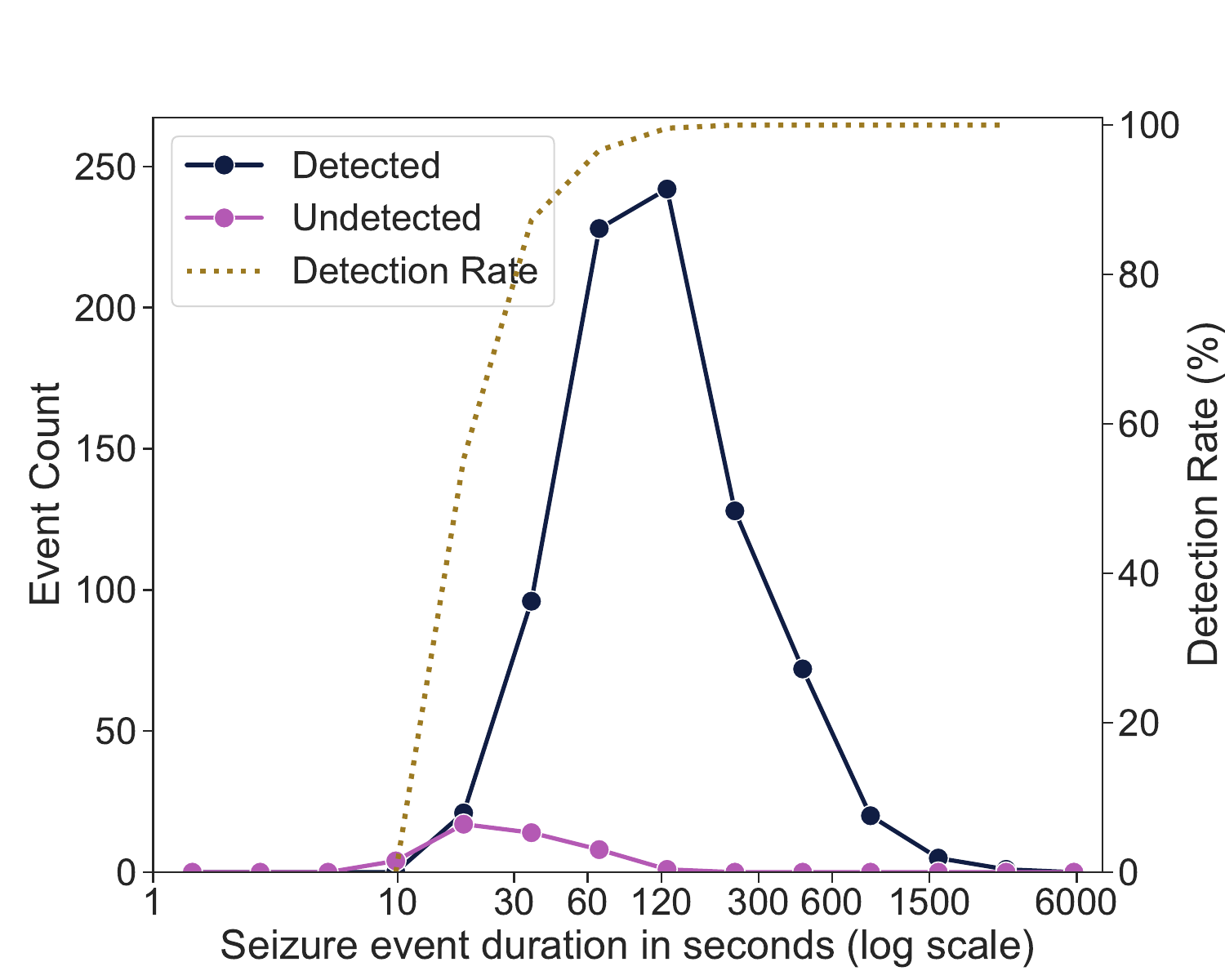}
    \end{subfigure}
    \begin{subfigure}[b]{0.45\textwidth}
        \includegraphics[width=\textwidth]{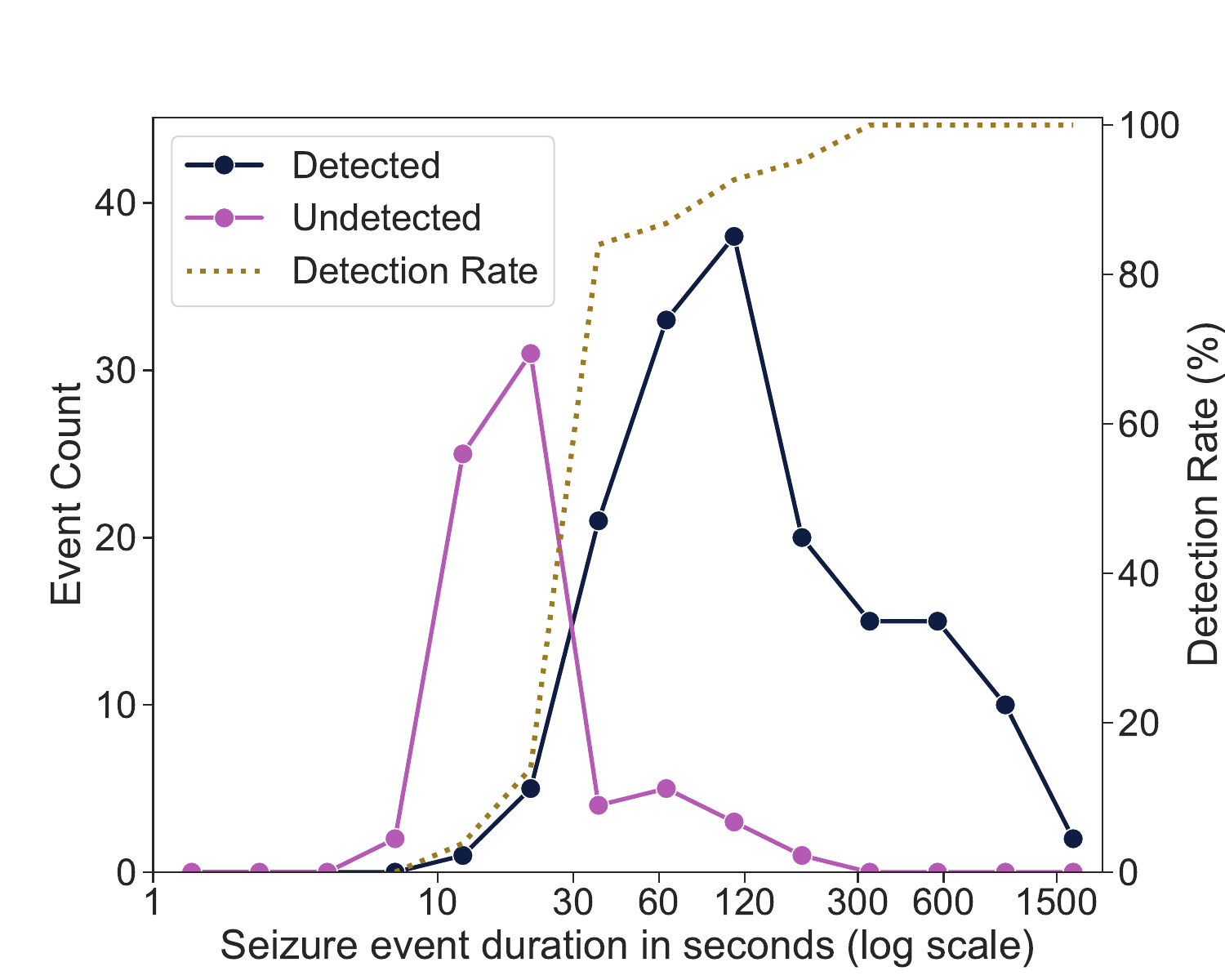}
    \end{subfigure}
    \caption{\small Extra-Large (XL) model performance by event duration of consensus seizures for
      Cork (left) and Helsinki (left) validation datasets. A notable finding is that most of the
      model errors are for short seizures ($<30$~s), where perhaps the 16~s input segment size
      limits detection resolution.}
    \label{fig:event_stats}
\end{figure}

Figure \ref{fig:event_stats} presents the distribution of model performance for increasing seizure-event durations. We find that for long seizures ($>$300~s) the model performs well, with a detection rate of 100\%.  Most of the missed events are for short seizures ($<$ 30~s). The difficulty with short seizures has a more pronounced effect on the Helsinki dataset where they were more commonly annotated. 

\subsection{Distribution shift}
\label{sec:distribution_shift}

Although we find strong scaling performance with model size for the Cork validation set, Fig.~\ref{fig:scaling_plots}c indicates diminishing returns for the Large and XL models on the Helsinki validation set. 
For those models, we find that an optimal classification threshold shifts down from 0.5 to 0.4 and 0.3 respectively. This may indicate that the larger, more capable models are learning some features that are useful on training sets but may not generalise to all settings. 

One hypothesis for why we observe this effect in the Helsinki dataset but not in our training data or the Cork validation set could be due to differences in clinical protocols applied in different centres. Indeed, as described in Section \ref{sec:dataset_details} there is a notable difference in the clinical demographics for the Helsinki dataset when compared to others. 

The most obvious difference is that almost 50\% ($38 / 78$) of the neonates have had EEG recorded $\ge$ 1 week after birth, in contrast to the Cork validation set which were all within a week of birth. If we divide the Helsinki dataset into two groups, those with EEGs recorded within a week (early-EEG group) and those with EEG recorded after the first week of life (late-EEG group), we find significant differences in primary diagnosis. A primary diagnosis of either asphyxia (including HIE) or stroke accounts for 92\% ($32/37$) in the early-EEG group compared with just 32\% ($10/31$) in the late-EEG group, $p<0.001$ ($n=68$; Fisher exact test). This may not be unexpected as the suspected diagnosis would likely be the main driver for EEG monitoring. 

\begin{figure}
    \centering
    \includegraphics[width=0.78\linewidth]{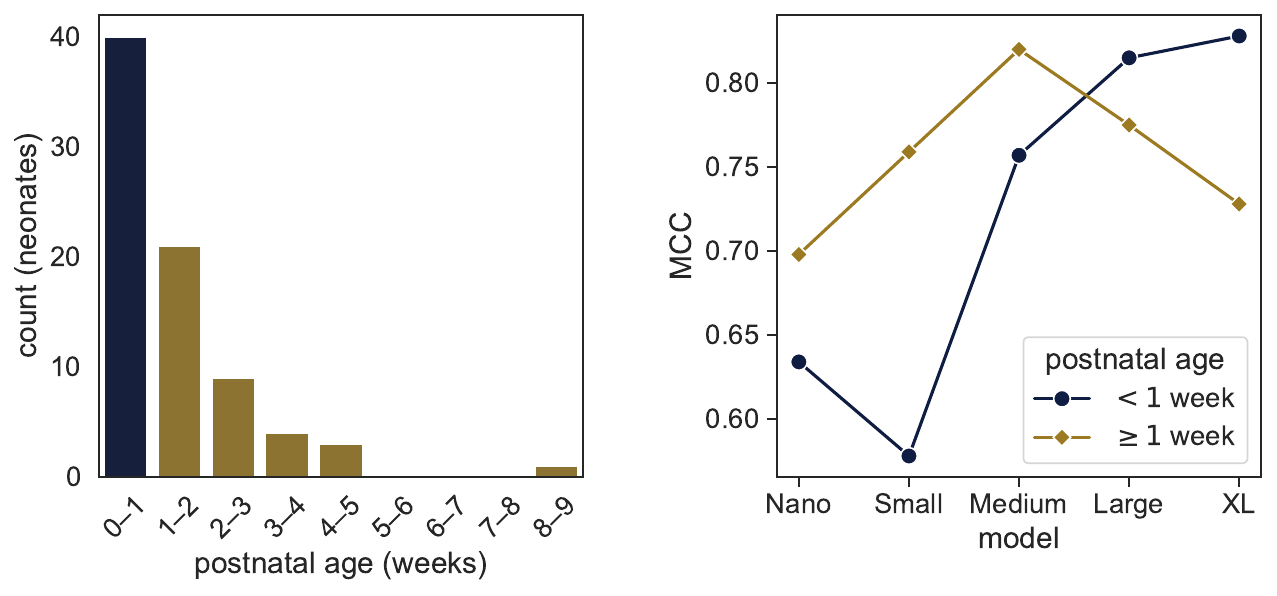}
    \caption{\small Divergence of scaling behaviour between 2 groups in the Helsink validation dataset. 
      Left: distribution of postnatal age in weeks. Right: Matthews correlation coefficient (MCC) for both groups.
      The scaling for $<1$ week postnatal age tracks closely with that observed
      in both Cork datasets, even matching the double-descent dip for the Small model. At $\geq1$ 
      week however we see progressive degradation for the Large and Extra-large (XL) models comparative to the
      Medium model.}
    \label{fig:postnatal-age}
\end{figure}

With this division of the dataset, we find a remarkable concordance between this explanation of the distribution shift and the scaling behaviour for these cohorts. In Fig.~\ref{fig:postnatal-age} we show that for the early-EEG group the same scaling behaviour observed in both Cork datasets is recovered. In contrast, for late-EEG group, we see that the performance peaks at the Medium model and starts to degrade progressively for the Large and XL models. This is suggestive that these more capable models are indeed learning something specific about EEG, which may be related to the primary diagnosis or to postnatal age. 

\subsection{Montage robustness}

A feature of our seizure detection model is its independence to channel montage, both the number of channels and the type of montage. To investigate this robustness, we take our predictions on the Helsinki dataset and simulate data loss or montage changes by randomly inserting contiguous sections of zeros in the per-channel model output. The final prediction is still calculated as the maximum over all channels so this dropped data will not contribute to the global estimate. We drop 10\%, 25\%, 50\%, and 100\% of the channel data at random in contiguous segments; here 100\% is equivalent to dropping the channel. This was applied across increasing numbers of channels until all but 1 were affected. This procedure was repeated for 20 trials.

\begin{figure}
    \centering
    \includegraphics[width=0.8\linewidth]{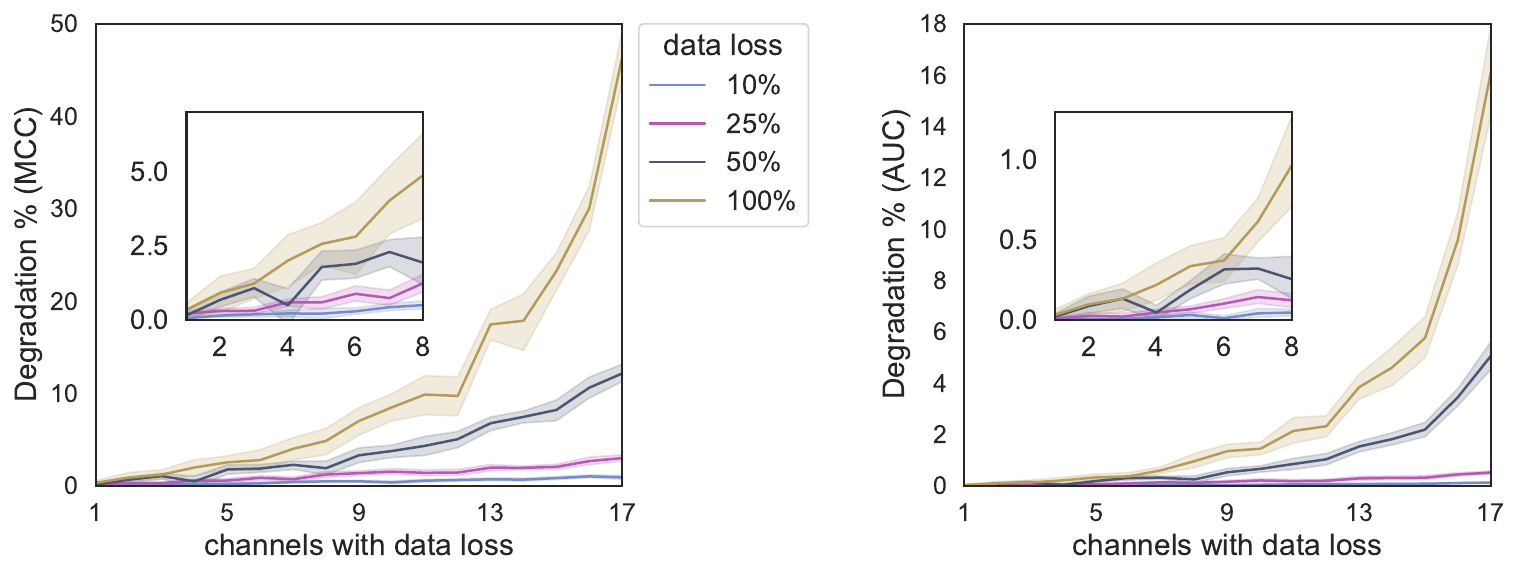}
    \caption{\small Summary of the effect of data loss on model performance on the Helsinki dataset. Degradation is measured relative to 0 data loss. Insert figures illustrate data loss for up to 8 channels.}
    \label{fig:montage-degradation}
\end{figure}

The result of this experiment is shown in Fig.~\ref{fig:montage-degradation}, where we summarise the impact as \% degradation relative to zero data loss for both the AUC and MCC metrics. We find that the model is remarkably robust: by dropping one-half of the channels the AUC (MCC) degrades by only 1.4\% (7.0\%). If the data loss is partial, we see even stronger results; for example, dropping 25\% from 17/18 channels we see only a 0.5\% (3.0\%) drop in AUC (MCC). The upper bound on performance here is of course determined by whether there is sufficient information remaining in the data to recover the global annotation even in principle, a dependence of the spatial distribution of the seizure event.

\end{appendices}

\bibliography{refs}

\end{document}